\documentclass[sigconf,natbib,screen=true]{acmart}
\AtBeginDocument{%
  \providecommand\BibTeX{{%
    \normalfont B\kern-0.5em{\scshape i\kern-0.25em b}\kern-0.8em\TeX}}}

\usepackage{enumitem}
\usepackage{multirow}
\usepackage{makecell}
\usepackage{caption}
\usepackage{subcaption,graphicx}
\usepackage{xspace}

\copyrightyear{2022} 
\acmYear{2022} 
\setcopyright{acmcopyright}\acmConference[WSDM '22]{Proceedings of the Fifteenth ACM International Conference on Web Search and Data Mining}{February 21--25, 2022}{Tempe, AZ, USA}
\acmBooktitle{Proceedings of the Fifteenth ACM International Conference on Web Search and Data Mining (WSDM '22), February 21--25, 2022, Tempe, AZ, USA}
\acmPrice{15.00}
\acmDOI{10.1145/3488560.3498440}
\acmISBN{978-1-4503-9132-0/22/02}

\newcommand{\partitle}[1]{\vspace{1mm}\noindent\textbf{#1}}
\newcommand{\ourMethod}{\emph{USi}\xspace}

\begin{document}

\fancyhead[]{}

\title[Evaluating Mixed-initiative Conversational Search Systems via User Simulation]{Evaluating Mixed-initiative Conversational Search Systems \\ via User Simulation}

\author{Ivan Sekuli\'c}
\affiliation{%
  \institution{Università della Svizzera italiana}
  \country{Lugano, Switzerland}
}
\email{ivan.sekulic@usi.ch}

\author{Mohammad Aliannejadi}
\affiliation{%
  \institution{University of Amsterdam}
   \country{Amsterdam, The Netherlands}
}
  \email{m.aliannejadi@uva.nl}

\author{Fabio Crestani}

\affiliation{%
  \institution{Università della Svizzera italiana}
   \country{Lugano, Switzerland}
}
\email{fabio.crestani@usi.ch}


\begin{abstract}
Clarifying the underlying user information need by asking clarifying questions is an important feature of modern conversational search system.
However, evaluation of such systems through answering prompted clarifying questions requires significant human effort, which can be time-consuming and expensive.
In this paper, we propose a conversational User Simulator, called \ourMethod, for automatic evaluation of such conversational search systems.
Given a description of an information need, \ourMethod is capable of automatically answering clarifying questions about the topic throughout the search session.
Through a set of experiments, including automated natural language generation metrics and crowdsourcing studies, we show that responses generated by \ourMethod are both inline with the underlying information need and comparable to human-generated answers.
Moreover, we make the first steps towards multi-turn interactions, where conversational search systems asks multiple questions to the (simulated) user with a goal of clarifying the user need.
To this end, we expand on currently available datasets for studying clarifying questions, i.e., Qulac and ClariQ, by performing a crowdsourcing-based multi-turn data acquisition. 
We show that our generative, GPT2-based model, is capable of providing accurate and natural answers to unseen clarifying questions in the single-turn setting and discuss capabilities of our model in the multi-turn setting.
We provide the code, data, and the pre-trained model to be used for further research on the topic.\footnote{https://github.com/isekulic/USi}
\end{abstract}





\maketitle

\section{Introduction}

The primary goal of a conversational search system is to satisfy user's information need by retrieving relevant information from a given collection.
In order to successfully do so, the system needs to have a clear understanding of the underlying user need.
Since user's queries are often under-specified and vague, a mixed-initiative paradigm of conversational search allows the system to take initiative of the conversation and ask the user clarifying questions, or issue other requests.
Clarifying the user information need has been shown beneficial to both the user and the conversational search system \cite{kiesel2018toward, aliannejadi2019asking, zamani2020generating}, providing a strong motivation for such mixed-initiative systems.


However, evaluation of the described mixed-initiative conversational search systems is not straightforward. 
The problem arises from the fact that expensive and time-consuming human-in-the-loop and human evaluation are required to properly evaluate conversational systems.
Such studies require real users to interact with the search system for several conversational turns and provide answers to potential clarifying questions prompted by the system.
A relatively simple solution is to conduct offline corpus-based evaluation \cite{aliannejadi2019asking}.
However, this limits the system to selecting clarifying questions from a pre-defined set of questions, which does not transfer well to the real-world scenario.
Moreover, such offline evaluation remains limited to single-turn interaction, as the pre-defined questions are associated with corresponding answers and are not aware of any previous interactions.
User simulation has been proposed to tackle the shortcomings of corpus-based and user-based evaluation methodologies.
The aim of a simulated user is to capture the behaviour of a real user, i.e., being capable of having multi-turn interactions on unseen data, while still being scalable and inexpensive like other offline evaluation methods~\cite{salle2021studying, zhang2020evaluating}.

In this paper, we propose a conversational User Simulator, called \ourMethod~ -- a model capable of multi-turn interactions with a general mixed-initiative conversational search system. 
Given an initial information need, \ourMethod interacts with the conversational system by accurately answering clarifying questions prompted by the system.
The answers are in line with the underlying information need and help elucidate the given intent.
Moreover, \ourMethod generate answers in fluent and coherent natural language, making its responses comparable to real users.
Previous work on the topic remained limited to retrieving answers from a pre-defined pool of human-generated answers to clarifying questions, e.g., CoSearcher \cite{salle2021studying}, or providing feedback with template-based answers in recommender systems~\cite{zhang2020evaluating}.

We base our proposed user simulator on a large-scale transformer-based language model, namely GPT-2~\cite{radford2019language}, ensuring the near-human quality of generated text.
Moreover, \ourMethod generates answers to clarifying questions in line with the initial information need, simulating the behaviour of a real user. 
We ensure that through a specific training procedure, resulting in a semantically-controlled language model.
We evaluate the feasibility of our approach with an exhaustive set of experiments, including automated metrics as well as human judgements.
First, we compare the quality of the answers generated by \ourMethod and several competitive sequence-to-sequence baselines by computing a number of automated natural language generation metrics.
\ourMethod significantly outperforms the baselines, which supports our decision to base our model on pre-trained GPT-2.
Second, we conduct a crowdsourcing study to assess how natural and accurate the generated answers are, compared to answers generated by humans.
The crowdsourcing judgements show no significant difference in the \emph{naturalness} and \emph{usefulness} of generated and human responses.
Third, we perform document retrieval evaluation following \cite{aliannejadi2020convai3,DBLP:conf/emnlp/AliannejadiKC0B21}, where retrieval is performed before and after answering a clarifying question.
We observe improvements in the retrieval performance when the answer is provided to the retrieval model, matching the retrieval performance of human-generated answers.
Next, we make the first steps towards multi-turn variant of the proposed model and present the multi-turn interaction dataset, acquired through crowdsourcing studies.
Finally, we discuss the intended use of the framework and demonstrate it's feasibility through a qualitative case study.

Our contributions can be summarised as follows:
\begin{itemize}[leftmargin=*]
    \item We propose a user simulator, \ourMethod, for conversational search system evaluation, capable of answering clarifying questions prompted by the search system. We release the code and pre-trained \ourMethod for future research.
    \item We perform extensive set of experiments to evaluate the feasibility of substituting real users with the user simulator, shedding new light on the upper bound of large-scale language models for the task. 
    \item We release a dataset of multi-turn interactions acquired through crowdsourcing, that we use to train our multi-turn version of the model. The dataset consists of $1000$ conversations of up to three turns, where crowdsourcing workers played the roles of the system that asks clarifying questions and the user seeking information.
\end{itemize}




\vspace{-1mm}
\section{Related Work}
Our work is part of a broad area of conversational information retrieval and user simulation. In this Section, we evaluate relevant work on the topics.

\partitle{Conversational search.}
Recent advancements of conversational agents have stimulated research in various aspects of conversational information access~\cite{ DBLP:conf/sigir/SunZ18,DBLP:conf/sigir/YanSW16,chuklin2019using}.
In fact, the report from the Dagstuhl Seminar N.\,19461 \cite{anand2020conversational} identifies conversational search as one of the essential areas of information retrieval (IR) in the upcoming years.
Moreover, \citet{radlinski2017theoretical} propose a theoretical framework for conversational search, highlighting the multi-turn user-system interactions as one of the desirable properties of modern conversational search.
This property is tied with a mixed-initiative paradigm in IR \cite{horvitz1999principles}, where the system is not only passive, but prompt the user with engaging content, such as clarifying questions.

Clarification has attracted considerable attention of the research community, including studies on human-generated dialogues on question answering (QA) forums, utterance intent analysis, and asking clarifying questions~\cite{DBLP:conf/chiir/BraslavskiSAD17}.
Asking clarifying questions has been shown to be beneficial for the conversational search system and the user.
E.g., \citet{kiesel2018toward} studied the impact of voice query clarification on user satisfaction and found that users like to  be prompted for clarification.
Moreover, \citet{aliannejadi2019asking} proposed an offline evaluation methodology for asking clarifying questions and showed the benefits of clarification in terms of improved performance in document retrieval once question is answered.
\citet{hashemi2020guided} proposed a Guided Transformer model for document retrieval and next clarifying question selection in a conversational search setting.
Furthermore, \citet{zamani2020generating} proposed reinforcement learning-based models for generating clarifying questions and the corresponding candidate answers from weak supervision data.
Moreover, \citet{sekulic2021towards} proposed GPT-2 based model for generating facet-driven clarifying questions.
Although extensive work related to clarification in search exists, effective and efficient evaluation methodologies of mixed-initiative approaches are scarce.

Another research direction in conversational search area is multi-turn passage retrieval, lead by the TREC Conversational Assistant Track (CAsT)~\cite{dalton2020trec}.
The system needs to understand the conversational context and retrieve appropriate passages from the collection.
As the further improvement, Ren et al.~\cite{ren2020conversations} introduced the task of conversations with search engines, where system generates a short, summarised response of the retrieved passages.
Other studies in the area of conversational search include user intent classification~\cite{qu2019user}, response ranking~\cite{dalton2020trec,sekulic2020extending,sekulic2020longformer}, document features for clarifying questions~\cite{sekulic2022exploiting}, user engagement prediction~\cite{sekulic2021user,DBLP:journals/corr/abs-2103-06192}, and query rewriting~\cite{peng2020few,vakulenko2021comparison,sekulic2022exploiting}.


In the field of natural language processing (NLP), researchers have studies question ranking~\cite{DBLP:conf/acl/DaumeR18} and generation~\cite{DBLP:journals/corr/abs-1904-02281,DBLP:conf/acl/HuangNWL18} in dialogue. These studies usually rely on large amount of data from query logs~\cite{DBLP:conf/www/RenNMK18}, industrial chatbots~\cite{DBLP:conf/acl/HuangNWL18}, and QA websites~\cite{DBLP:conf/acl/DaumeR18,DBLP:journals/corr/abs-1904-02281,DBLP:conf/acl/TianYMSFZ17}. 
For example, \citet{DBLP:conf/acl/DaumeR18} developed a neural model for question selection on an artificial dataset of clarifying questions and answers extracted from QA forums. In their later study, they proposed an adversarial training mechanism for generating clarifying questions given a product description from Amazon~\cite{DBLP:journals/corr/abs-1904-02281}. 
Unlike these studies, we study user-system interaction in an IR setting, where the user's information need is presented in the form of short queries (vs.~a long detailed post on StackOverflow), with a result of ranked list of relevant documents. 
Furthermore, the IR system can ask clarifying question to elucidate user's information need, which then needs to be answered.

\partitle{User simulation in information retrieval.}
Given the complexity of human-computer interactions and natural language, there has been an ongoing discussion in the NLP community about the credibility of automatic evaluation metrics that are based on text overlap~\cite{novikova2017we}. Metrics such as BLEU and ROGUE that try to judge a system's output solely based on how much overlap it has with a reference utterance cannot capture the performance of the system accurately~\cite{belz2006comparing}. Hence, human annotation should be done to evaluate a system's performance when a generative model is used, in tasks such as summarisation and machine translation. Moreover, evaluation of a system becomes even more complex if an ongoing interaction between the user and system exists. Not only must the system evaluate the generated utterance, it should also be able to incorporate a human response. For this reason, researchers adopt  human-in-the-loop techniques to mimic human-computer interactions, and further perform human annotation to evaluate the whole system's performance (in response to human). Recent work of \citet{lipani2021doing} propose a metric for offline evaluation of conversational search systems based on user interaction model. 

With an idea to alleviate the need for time-consuming and expensive human evaluation, researchers proposed replacing the user with a user simulation system~\cite{salle2021studying,sun2021simulating}.
Simulation in IR has long been studied (1973) \cite{cooper1973simulation} with the idea of generating pseudo-docs and pseudo-queries to study literature search system performance. The work was then followed by \citet{griffiths1976computer}, proposing a general framework of simulation for IR systems. \citet{tague1980problems} later studied the problems for user simulation in bibliographic retrieval systems. User simulation for evaluation was first proposed in 1990 by \citet{gordon1990evaluating} where the authors proposed a framework for generating simulated queries. This work has been long followed in the literature to study various hypothetical user and system actions (e.g., issuing 100 queries in a session) that cannot be done in a real system~\cite{DBLP:conf/sigir/Azzopardi11}. In particular, \citet{DBLP:conf/sigir/Azzopardi11} proposed to study the cost and gain of user and system actions and studied the effect of different strategies using simulated queries and actions of users (e.g., clicking on relevant documents). \citet{mostafa2003simulation} studied different dimensions of users' interests and their impact on user modelling and information filtering. \citet{diaz2009adaptation} adapted an offline vertical selection prediction model in the presence of user feedback for user simulation.

More recently, there has been research on simulating users to evaluate the effectiveness of systems~\cite{carterette2011simulating,sun2021simulating,salle2021studying,zhang2020evaluating,yang2016trec}. \citet{carterette2011simulating} proposed a conceptual framework for investigating various aspects of simulations, namely, system effectiveness, user models, and user utility. With the recent developments of conversational systems, more attention towards simulating users in a conversation has been drawn. \citet{sun2021simulating} proposed a simulated user for evaluating conversational recommender systems based on predefined actions and structured response types. \citet{salle2021studying} proposed a parametric user simulator for information-seeking conversation where the simulator takes an information need and responds to the system accordingly. In fact, this work is the closest work to ours. However, we would like to draw attention to various limitations of this work. Even though this work takes an information need as input and aims at answering to the system's request according to that, it fails to generate responses. The approach is limited to predicting the relevance of the system's utterance to the user's information need and selecting an appropriate answer from a list of human-generated answers. In this work, we take one step further and generate human-like answers in natural language. Also, the work by \citet{zhang2020evaluating} that simulates users for recommender system evaluation, uses structured data and response types. In this work, we propose a simulator that generates natural language responses based on unstructured data.










\vspace{-1mm }
\section{Simulated user}
\label{sec:usi}
In this section, we explain the role of a user in the evaluation of conversational search and dialogue systems.
Next, we define several characteristics a user simulator should have in order to be able to replace real users in certain evaluation tasks.

\vspace{-1mm}
\subsection{User's role in evaluating conversational search systems}
Previous work in task-oriented dialogue systems and conversational search systems mostly evaluate the performance of the systems in an offline setting using a corpus-based approach \cite{deriu2021survey}.
However, offline evaluation does not accurately reflect the nature of conversational systems, as the evaluation is possible only at a single-turn level.
Thus, in order to properly capture the nature of the conversational search task,  it is necessary to involve users in the evaluation procedure \cite{black2011spoken,li2019acute}.
User involvement allows proper evaluation of multi-turn conversational systems, where user and system take turns in a conversation.
Nonetheless, while such approach most precisely captures the performance of the systems in a real-world scenario, it is tiresome, expensive, and unscalable.
In pursuit of alleviating the evaluation of dialogue systems, while still accurately capturing the overall performance, a simulated user approach has been proposed \cite{zhang2020evaluating, sun2021simulating}.
The simulated user is intended to provide a substitute for real users, as it is easily scalable, cheap, fast, and consistent.
Next, we formally describe the characteristics of a simulated user for conversational search system evaluation.

\vspace{-1mm}
\subsection{Problem definition}
As already mentioned, evaluating conversational search systems is hard due to the necessity of human judgements at each turn of an interaction with the search system.
In this work, we aim to alleviate the procedure of evaluating certain types of mixed-initiative conversational search systems.
Specifically, we provide a simulated user with an imaginary information need capable of answering various types of clarifying questions prompted by any modern conversational search system.

Formally, our simulated user $U$ is initialised with a given information need $in$. 
Simulated user $U$ formulates its need in a form of the initial query $q$, which is then given to the general mixed-initiative conversational system $S$.
The aim of the system $S$ is to elucidate the information need $in$ through a series of clarifying questions $cq$.
We do not go into details of the implementation of such a system, but different approaches have been proposed in recent literature \cite{aliannejadi2019asking, hashemi2020guided}.
Next, the simulated user $U$ needs to provide an answer $a$ to the system's question.
The answer $a$ needs to be in line with user's information need $in$.

\partitle{Single-turn responses.}
Formally, user $U$ needs to generate an answer $a$ to the system's clarifying question $cq$, conditioned on the initial query $q$ and the original user's intent $in$:
\vspace{-1mm}
\begin{equation}
\label{eq:user1}
    a = f(cq | in, q)
\vspace{-1mm}
\end{equation}
The user $U$ is expected to answer the question in line with its information need, not just based on a potentially vague and under-specified query, like traditional chatbots would be inclined to do.

\partitle{Conversation history-aware user.}
Moreover, the system can take further initiative and ask additional clarifying questions. 
Thus, our simulated user $U$ needs to track the conversation flow as well.
Formally, at the conversational turn $i$, $U$ generates an answer given by:
\vspace{-1mm}
\begin{equation}
\label{eq:user2}
    a_i = f(cq_i | in, q, H)
\end{equation}
\noindent where $H$ is conversational history, consisting the interaction between the user and the system up until the current turn: $H = \{(cq_j, a_j)\}$, where $j \in [1\dots i-1]$. 
In the next Section, we explain how we modelled the described simulated user.

\section{Simulation Methodology}
\label{sec:methodology}
In this section, we motivate and describe in detail our proposed User Simulator, \ourMethod.
We make \ourMethod semantically-controlled through specific language modelling training.
We base our simulated user on a large-scale transformer-based model, namely GPT-2 \cite{radford2019language}.

\subsection{Semantically-controlled text generation}
We define the task of generating answers to clarifying questions as a sequence generation task. 
Thus, we employ language modelling as our main tool for generating sequences.
The goal of a language model (LM) is to learn the probability distribution $p_\theta(x)$ of a sequence of length $n$: $x = [x_1, x_2,\dots,x_n]$, where $\theta$ are the parameters of the LM.
Current state-of-the-art language models, such as GPT-2, learn the distribution in an auto-regressive manner, i.e., formulating the task as next-word prediction task:
\vspace{-1mm}
\begin{equation}
\label{eq:lm}
p_\theta(\boldsymbol{x}) = \prod_{i=1}^{n}p_\theta(x_i|x_{<i})
\vspace{-1mm}
\end{equation}

However, recent research showed that large-scale transformer-based language models, although generating text of near-human quality, are prone to ``hallucination'' \cite{dziri2021neural} and in general lack semantic guidance \cite{rosset2020leading}.
Thus, with a specific fine-tuning technique and careful input arrangement, we fine-tune semantically-conditioned LM.
As mentioned in the previous section, answer generation needs to be conditioned on the underlying information need. 
To this aim, we learn the probability distribution of generating an answer $a$:
\begin{equation}
\label{eq:conditioned}
p_\theta(\boldsymbol{a}|in,q,cq) = \prod_{i=1}^{n}p_\theta(a_i|a_{<i},in,q,cq)
\end{equation}
\noindent where $a_i$ is the current token of the answer, $a_{<i}$ are all the previous ones, while $in$, $q$, and $cq$ correspond to the information need, the initial query, and the current clarifying question from Equation \ref{eq:user1}, respectively.

\subsection{GPT2-based simulated user}
GPT-2 is a large-scale transformer-based language model trained on a dataset of 8 million web pages, capable of synthesising text of near human quality \cite{radford2019language}.
Moreover, as it is trained on an extremely diverse dataset, it can generate text on various topics, which can be primed with an input sequence.
GPT-2 has previously been used for various text generation tasks, including dialogue systems and chatbots \cite{budzianowski2019hello}.
Therefore, it is a suitable choice for our task of simulating users through generating answers to clarifying question in a conversational search system.

We base our proposed user simulator \ourMethod on the GPT-2 model with language modelling and classification losses, i.e., DoubleHead GPT-2. In this variant, the model not only learns to generate the appropriate sequence through the language modelling loss, but also how to distinguish a correct answer to the distractor one. This has been shown to improve the sequence generation \cite{radford2019language} and has showed superior performance over only-language loss GPT-2 in the initial stage of experiments. The two losses are linearly combined.

\partitle{Single-turn responses.}
We formulate the input to the GPT-2 model, based on Equation \ref{eq:conditioned}, as:
\vspace{-1mm}
\begin{equation}
\label{eq:gptinput}
    input\_seq = in [SEP] q [SEP] cq [bos] a [eos]
\vspace{-1mm}
\end{equation}

\noindent where $[bos]$, $[eos]$, and $[SEP]$ are special tokens indicating the beginning of sequence, the end of sequence, and a separation token, respectively.
Information need $in$, initial query $q$, clarifying question $cq$, and a target answer $a$ are tokenized prior to constructing the full input sequence to the model. 
Additionally, we construct segment embeddings, which indicate different segments of the input sequence, namely $in$, $q$, $cq$, and $a$. 

When training the DoubleHead variation of the model, we formulate the first part of the input as described above. Additionally, we sample the ClariQ dataset for distractor answers and process them in the same manner as the original answer, based on Equation \ref{eq:gptinput}.
Therefore, the DoubleHead GPT-2 variant accepts as input two sequences, one with the original target answer in the end, and the other with the distractor answer. 
It then needs to not only learn to model the target answer, but also to distinguish between original and distractor answers and provide a binary label indicating which of the two answers is the desirable one.
We sample the distractor answers from the aforementioned datasets. When possible, we ensure that if the target answer starts with ``Yes'', the distractor answers starts with ``No'', in order to enforce the connection between the answer, the clarifying question, and the information need.
Likewise, if the answer starts with ``No'', we sample a distractor answer that starts with ``Yes''.
Note that \ourMethod does not generate answers that begin strictly with a ``yes'' or a ``no''.

\partitle{Conversation history-aware model.}
The conversation history-aware model calls for a different formulation of the input and the training.
Specifically, the input to history-aware GPT-2 is constructed as:
\vspace{-1mm}
\begin{equation*}
    \label{eq:gptinput2}
    \resizebox{\columnwidth}{!}{%
    $input\_seq = in [user] q [system] cq_{<i} [user] a_{<i} [system] cq_i [bos] a_i [eos]$
    }%
    \vspace{-1mm}
\end{equation*}
\noindent where $[user]$ and $[system]$ are additional special tokens indicating the conversational turns between the (simulated) user and the conversational system, respectively.

\partitle{Inference.}
During inference, we omit the answer $a$ from the input sequence, as our goal is to generate this answer to a previously unseen question.
In order to generate answers, we use a combination of state-of-the-art sampling techniques to generate a textual sequence from the trained model.
Namely, we utilise temperature-controlled stochastic sampling with top-$k$ \cite{fan2018hierarchical} and top-$p$ (nucleus) filtering \cite{holtzman2019curious}.
After some initial experiments and consultation with previous work, we fix the parameters of the temperature to $0.7$, $k$ to $0$, and $p$ to $0.9$.


\vspace{-1mm}
\section{Data}
\label{sec:data}

\subsection{Qulac and ClariQ}

\begin{table}[t]
\vspace{-7mm}
\caption{Statistics for Qulac and ClariQ datasets.}
\vspace{-3mm}
\label{tbl:stats}
\begin{tabular}{lll}
\toprule
 & Qulac & ClariQ \\
 \midrule
Number of topics & 198 & 237 \\
Number of facets & 762 & 891 \\
Number of questions & 2,639 & 3,304 \\
Number of question-answer pairs & 10,277 & 11,489 \\
\bottomrule
\end{tabular}
\vspace{-3mm}
\end{table}

For the purpose of training and evaluating our proposed simulated user \ourMethod, we utilise two publicly available datasets, Qulac \cite{aliannejadi2019asking} and ClariQ \cite{aliannejadi2020convai3}.
The aim of both datasets is to foster research in the field of asking clarifying questions in open-domain conversational search.
Qulac was created on top of the TREC Web Track 2009-12 collection. The Web Track collection contains ambiguous and faceted queries, which often require clarification when addressed in a conversational setting. 
Given a topic from the dataset, clarifying questions were collected via crowdsourcing.
Then, given a topic and a specific facet of the topic, workers were employed to gather answers to these clarifying questions.
This results in a tuple of ($topic$, $facet$, $clarifying\_question$, $answer$).
Most of the topics in the dataset are multi-faceted and ambiguous, meaning that the clarifying questions and answers need to be in line with the actual facet.
ClariQ is an extension of Qulac created for the ConvAI3 challenge \cite{aliannejadi2020convai3} and contains additional non-ambiguous topics.
Relevant statistics of the datasets are presented in Table \ref{tbl:stats}.

We utilise these datasets by feeding the corresponding elements to Equation \ref{eq:conditioned}.
Specifically, \emph{facet} from Qulac and ClariQ represents the underlying information need, as it describes in detail what the intent behind the issued \emph{query} is. 
Moreover, \emph{question} represents the current asked question, while \emph{answer} is our language modelling target.

\subsection{Multi-turn conversational data}
A major drawback of Qulac and ClariQ is that they are both built for single-turn offline evaluation.
In reality, a conversational search system is likely to engage in a multi-turn dialog in order to elucidate user need.
To bridge the gap between single- and multi-turn interactions, we construct multi-turn data that resembles a more realistic interaction between a user and the system.
Our user simulator \ourMethod is then further fine-tuned on this data.

To acquire the multi-turn data, we construct a crowdsourcing-based human-to-human interaction.
At each conversational turn, crowdsourcing worker is tasked to behave as a search system by asking a clarifying question on the topic of the conversation.
Then, another worker is tasked to provide the answer to that question having in mind the underlying information need and the conversation history, imitating the behaviour of the real user.
We construct in $500$ conversations up to depth of three, i.e., we have three sequential question-answers pairs for a topic and its facet.

In order to further study the effects certain clarifying questions have on the whole search experience, we construct several edge cases.
In such cases, the clarifying question prompted by the search system is considered faulty, as it is either a repetition, off topic, unnecessary, or completely ignores previous user's answers. 
We obtain answers to these questions to provide a more realistic data for the training of our model, making our simulated user as human-like as possible. 
These clarifying questions are intended to simulate a conversational search system of poor quality and provide insight into user's responses to such questions.
We employ workers to provide answers to additional $500$ clarifying questions of poor quality, up to the depth of two.
The specific edge cases and their descriptions with examples are presented in Table \ref{tbl:human_data}. We publicly release the acquired multi-turn datasets.


\begin{table}[t]
\vspace{-7mm}
\caption{Multi-turn dataset acquired through crowdsourcing for fine-tuning a more realistic user simulator. Sample conversations of depth 3 are omitted for space purposes.}
\vspace{-3mm}
\label{tbl:human_data}
    \centering
    \resizebox{\columnwidth}{!}{%
    \begin{tabular}{llll}
    \toprule
         Question case & Description & Sample conversation & N  \\
         \midrule
         Normal & \makecell[tl]{A good system naturally\\ continues the conversation.} & \makecell[tl]{U: I'm looking for information on dieting\\S: Are you looking for dieting tips? \\U: Yes and exercise tips as well\\S: Do you need anything specific in\\ relation to counting calories\\ you consume daily?\\U: Yes, I would like to know more\\ about that topic.} & 500 \\\Xhline{0.2\arrayrulewidth}
         
         Repeat & \makecell[tl]{System repeats the \\previous question.} & \makecell[tl]{U: Find information on raised gardens.\\ S: Do you need information on\\ materials needed? \\ U: No, I want to find plans \\S: Do you need information on \\ materials needed?\\U: I want what I previously asked for. } & 50 \\\Xhline{0.2\arrayrulewidth}
         
         Off-topic & \makecell[tl]{System asks the user an \\ off-topic question.} & \makecell[tl]{U: I'm looking for an online world atlas. \\S: Are you interested in satellite maps?\\ U: No, I want an online world atlas\\S: Which mountain ski resort would you\\ like information around the pocono area? \\U: I am not interested in this topic.} & 50 \\\Xhline{0.2\arrayrulewidth}
         
         Similar & \makecell[tl]{System asks a question\\ similar to the previous one,\\ ignoring the user's answer.} & \makecell[tl]{U: I'm looking for information about \\ mayo clinic Jacksonville FL\\ S: Would you like to request\\ an appointment?\\ U: yes \\S: Are you looking for the address\\ of mayo clinic jacksonville fl?\\U: I just want to request an appointment.} & 400 \\
         \bottomrule
    \end{tabular}
    }
    \vspace{-3mm}
\end{table}


\begin{table*}[t]
\vspace{-7mm}
\caption{Performance by automated NLG metrics on Qulac (25 topics) and ClariQ (dev set).}
\vspace{-3mm}
\label{tbl:nlg}
    \begin{tabular}{ll|llllll}
    \toprule
     & model & BLEU-1 & BLEU-2 & BLEU-3 & ROUGE\_L & SkipThoughtCS & EmbeddingAvgCS  \\
     \midrule
    \multirow{3}{*}{\rotatebox[origin=c]{90}{Qulac}} & LSTM-seq2seq & 0.1993 & 0.1446 & 0.1076 & 0.2428 & 0.3091 & 0.7468  \\
     & Transformer-seq2seq & 0.2071 & 0.1317 & 0.0886 & 0.1997 & 0.3118 & 0.6566  \\
     & \ourMethod & \textbf{0.2495} & \textbf{0.1595} & \textbf{0.1079} & \textbf{0.2495} & \textbf{0.4167} & \textbf{0.7896}  \\[0.5ex] 
    \multirow{3}{*}{\rotatebox[origin=c]{90}{ClariQ}} & LSTM-seq2seq & 0.1989 & 0.1401 & 0.0988 & 0.2210 & 0.3158 & 0.7012 \\
     & Transformer-seq2seq & 0.2041 & 0.1352 & 0.0936 & 0.2067 & 0.3666 & 0.7077  \\
     & \ourMethod & \textbf{0.3029} & \textbf{0.2404} & \textbf{0.2054} & \textbf{0.2359} & \textbf{0.4025} & \textbf{0.7322} \\
     \bottomrule
    \end{tabular}
\end{table*}

\section{Evaluation}
\label{sec:evaluation}

Our aim is to evaluate whether our proposed simulated user can replace real users in answering clarifying questions of conversational search systems, which would make the evaluation of such systems significantly less troublesome.
Overall, we aim to answer four main research questions:
\begin{description}
    \item [RQ1:] To what extent are the generated answers in line with the underlying information need?
    \item [RQ2:] How coherent and natural is the language of the generated answers?
    \item [RQ3:] To what extent does the retrieval model of the conversational search system benefit from the generated answers?
    \item [RQ4:] How does \ourMethod behave in multi-turn interactions?
\end{description}

To address these questions, we first compute several natural language generation metrics to compare the generated answers to the oracle human answers from ClariQ.
As several NLG metrics received criticism from the NLP community, especially since they do not correlate well with the coherence of the text, we perform a crowdsourcing study to evaluate the \emph{naturalness} of generated answers.
In order to evaluate whether the generated answers are in line with the actual information need, we carry out additional crowdsourcing study, evaluating the \emph{usefulness} of answers.
Moreover, we analyse the impact of generated answers to retrieval model performance,  by performing a document retrieval before and after answering the prompted clarifying question, as described in Section \ref{sec:retrieval}.
Finally, we perform qualitative analysis of generated answers.

We compare our GPT-2-based user simulator to two competitive sequence-to-sequence baselines. The first baseline is a multi-layer bidirectional LSTM encoder-decoder network for sequence-to-sequence tasks \cite{sutskever2014sequence}.\footnote{We use the IBM implementation for our experiments: \url{https://github.com/IBM/pytorch-seq2seq}}
The second baseline is a transformer-based encoder-decoder network, based on Vaswani et al.\,\cite{vaswani2017attention}.
We perform hyperparameter search to select the learning rate, number of layers, and hidden dimension of the models.
Both baselines are trained with the same input as our main model.

\subsection{Automated NLG metrics}
\label{sec:nlgeval}

We first study the language generation ability of \ourMethod and of the aforementioned baselines.
We compute several standard metrics for evaluating the generated language. 
We use two widely adopted metrics based on n-gram overlap between the generated and the reference text. These are BLEU \cite{papineni2002bleu} and ROUGE \cite{lin2004rouge}.
Next, we compute the EmbeddingAverage and SkipThought metrics aiming to capture the semantics of the generated text, as they are based on the word embeddings of each token in the generated and the target text. The metric is then defined as a cosine similarity between the means of the word embeddings in the two texts \cite{kryscinski2019neural}.
The models are trained on ClariQ training set and evaluated on unseen ClariQ development set. 
We evaluate on ClariQ's development set since the test set does not contain question-answer pairs.
We take a small portion of the training set for our actual development set.
The answers generated by \ourMethod and the baselines are compared against oracle answers from ClariQ, generated by humans.


\subsection{Response Naturalness and Usefulness}
\label{sec:naturaleval}
In order to simulate a real user, the generated responses by our model need to be fluent and coherent.
Thus, we study the \emph{naturalness} of the generated answers. We define \emph{naturalness} as an answer being natural, fluent, and likely generated by a human.
Similarly, fluency \cite{callison2006re} and humanness \cite{see2019makes} have been used for evaluating generated text.
Moreover, we assess the \emph{usefulness} of the answers generated by our simulated user. 
We define \emph{usefulness} as an answer being in line with the underlying information need and guiding the conversation towards the topic of the information need. 
This definition of usefulness can be related to similar metrics in previous work, such as adequacy \cite{stent2005evaluating} and informativeness \cite{chuklin2019using}.

We perform a crowdsourcing study to assess the \emph{naturalness} and \emph{usefulness} of generated answers to clarifying questions. 
We use Amazon MTurk for acquiring workers, based in US, with at least 95\% task approval rate.
The study was done in a pair-wise setting, i.e., each worker was presented with a number of answer pairs, where one of the answers was generated by our model and the other was generated by a human, taken from the ClariQ collection. 
Their task was then to provide judgement on which answers is more natural or useful, depending on the study. 
The workers have been provided with the context, i.e., the initial query, facet description, and clarifying question.

We annotate 230 answer pairs for \emph{naturalness} and 230 answer pairs for \emph{usefulness}, each judged by two crowdsource workers.
We define a \emph{win} for our model if both annotators voted our generated answer as more natural/useful, and \emph{loss} for our model if both voted the human generated answer as more natural/useful. In case the two workers voted differently on a single answer pair, we define that as a \emph{tie}.
With this study, we aim to shed light onto research questions RQ1 and RQ2, i.e., whether the generated answers are indeed natural and in line with the underlying information need, compared to human-generated answers.
Additionally, we compare Transformer-seq2seq to \ourMethod.
The results of the study are discussed in Section \ref{sec:natural}.

\vspace{-5mm}
\subsection{Impact of generated answers to document retrieval Performance}
\label{sec:retrieval}

As the basis for the offline evaluation of open-domain conversational search systems, Aliannejadi et al.\,\cite{aliannejadi2019asking} propose the document retrieval task with the answer to the prompted clarifying question.
The initial query is expanded with the text of the clarifying question and the user's answer and then fed to a retrieval model, such as BM25 or Query Likelihood.
The intuition is that if the clarifying question and the answer were both useful, the retrieval model will perform better with them in input, alongside the initial query. 
In fact, they show significant improvements in retrieval performance with the additional input compared to query-only setting, which is in general a strong motivation for asking clarifying questions in conversational search.
The initial retrieval is performed on ClueWeb09b collection, while queries are taken from the ClariQ development set.
Each query is associated with the information need (facet) description and several clarifying questions.
We then generate answers to these questions and perform additional document retrieval with the initial query expanded with generated answers and corresponding questions.

We follow the described evaluation paradigm to assess whether our simulated user generates useful answers, compared to the human-generated ones.
With this study, we aim to answer research question RQ3, i.e., how beneficial are the generated answers to the retrieval model of a conversational search system.
Our hypothesis is that the retrieval performance should increase when the initial query is expanded with the generated answers.
The results of the experiment are discussed in Section \ref{sec:retrievalresults}.


\section{Results and Discussion}
\label{sec:results}

\subsection{Automated NLG metrics}
\label{sec:nlgresults}


Performance of the baseline model and our simulated user, as evaluated by automated NLG metrics described in Section \ref{sec:nlgeval}, is presented in Table \ref{tbl:nlg}.
\ourMethod significantly outperforms all baselines by all of the computed metrics both on Qulac and ClariQ.
Even though LSTM-seq2seq showed strong performance in various sequence-to-sequence tasks, such as translation \cite{sutskever2014sequence} and dialogue generation \cite{shao2017generating},  it performs quite poorly on our task. 
Similar outcome is observed for Transformer-seq2seq.
We hypothesise that the poor performance in this task is due to limited training data, as the success of these seq2seq models on various different tasks was conditioned on large training sets.
Our GPT2-based model does not suffer from the same problem, as it has been pre-trained on a large body of text, making the fine-tuning enough to capture the essence of the task, which is generating answers to clarifying questions.

\subsection{Naturalness and Usefulness}
\label{sec:natural}

\begin{table}[]
    \centering
    \vspace{-7mm}
    \caption{Results on \emph{naturalness} and \emph{usefulness} of responses, \ourMethod vs human-generated..}
    \vspace{-3mm}
    \label{tbl:natural}
    \begin{tabular}{l|lll}
    \toprule
        & \ourMethod Wins & Human Wins & Ties \\
    \midrule
         \emph{Naturalness} & 17\% & 38\% & 45\% \\
         \emph{Usefulness} & 22\% & 27\% & 51\% \\
    \bottomrule
    \end{tabular}
    \vspace{-3mm}
\end{table}

Table \ref{tbl:natural} presents the results of the crowdsourcing study on \emph{usefulness} and \emph{naturalness}, comparing answers generated by \ourMethod and human, as described in Section \ref{sec:naturaleval}.
Both in terms of \emph{naturalness} and \emph{usefulness}, we observe a large number of \emph{ties}, i.e., the two workers annotating the answer pair did not agree which one is more natural/useful.
Since we are comparing answers generated by our GPT2-based simulated user with the answers written entirely by humans, this result goes in favour of our proposed model.
Moreover, the difference between \emph{losses} and \emph{wins} for our model is relatively small (38\% losses, 17\% wins) for \emph{naturalness}, and even smaller in terms of \emph{usefulness} (32\% losses, 23\% wins). 
We conduct trinomial test for statistical significance \cite{bian2011trinomial}, an alternative to the Sign and binomial tests that takes into account ties.\footnote{Another point-of-view would be to test for equivalent effectiveness \citet{jayasinghe2015statistical}, however, we refrain from it since it does not take ties into account.}
In terms of \emph{naturalness}, we reject the null hypothesis of equal performance with $p < 0.05$, i.e., human generated answers are more natural than ones generated by \ourMethod.
Nonetheless, $45\%$ of ties between \ourMethod- and human-generated answers suggests the high quality of the generated text.
Regarding \emph{usefulness}, we accept the null hypothesis of equal performance with $p = 0.43$, i.e., there is no statistically significant difference between the performance of human annotators and \ourMethod.

Table \ref{tbl:natural2} presents the results of the comparison between the Transformer-seq2seq and \ourMethod.
We observe a win of the proposed \ourMethod over the baseline by a large margin. Our GPT-2-based model significantly outperforms the baseline ($p < 0.05$) both in terms of \emph{naturalness} (50\% wins and 3\% losses) and \emph{usefulness} (66\% wins and 3\% losses).
This finding is in line with the automated evaluation of generated answers.

Regarding the research questions RQ1 and RQ2, i.e., whether the responses generated by our model are in line with the underlying information need and at the same time coherent and fluent, we arrive to the conclusion of satisfactory performance of the simulated user.
The generated answers to clarifying questions seem to be able to compete with the answers produced by humans both in terms of \emph{naturalness} and \emph{usefulness}.
Moreover, strong performance of \ourMethod over Transformer-seq2seq additionally motivates the use of large-scale pre-trained language models, such as GPT-2, for the task.
These results make a strong case for the utilisation of a user simulator for mixed-initiative conversational search system evaluation.

\begin{table}[]
    \centering
    \vspace{-7mm}
    \caption{Results on \emph{naturalness} and \emph{usefulness} of responses, \ourMethod vs Transformer-seq2seq baseline.}
    \vspace{-3mm}
    \label{tbl:natural2}
    \begin{tabular}{l|lll}
    \toprule
         & \ourMethod Wins & Baseline Wins & Ties \\
         \hline
         \emph{Naturalness} & 50\% & 3\% & 47\% \\
         \emph{Usefulness} & 66\% & 3\% & 31\% \\
    \bottomrule
    \end{tabular}
    \vspace{-3mm}
\end{table}

\vspace{-1mm}
\subsection{Document retrieval performance}
\label{sec:retrievalresults}

\begin{table*}[t]
\centering
\vspace{-7mm}
\caption{Document retrieval performance based on the answers provided by our simulated user. Percentages in parentheses report relative increase or decrease in performance over Oracle. Symbols $\dagger$ and $\ddagger$ indicate statistically significant difference compared to the query-only baseline and the human-generated answers, respectively. The significance is reported under two-sided t-test with $p<0.01$.}
\vspace{-3mm}
\label{tab:retrieval}
    \begin{tabular}{l|lllll}
    \toprule
                   & nDCG@1 & nDCG@5 & nDCG@20 & P@1 & MRR@100 \\
                   \midrule
    Query-only     & 0.1304 (-3\%) & 0.1043 (-21\%) & 0.0852 (-26\%) & 0.1764 (-4\%) & 0.2402 (-12\%) \\
    LSTM-seq2seq          & 0.1018$\ddagger$ (-24\%) & 0.0899$\ddagger$ (-31\%) & 0.0745$\ddagger$ (-35\%) & 0.1409$\ddagger$ (-23\%) & 0.2131$\ddagger$ (-22\%)\\
    Transformer-seq2seq        & 0.1124 (-16\%) & 0.1040$\ddagger$ (-21\%) & 0.0847$\ddagger$ (-26\%) & 0.1559$\ddagger$ (-15\%) & 0.2309$\ddagger$ (-15\%)\\
    \ourMethod        & 0.1355 (+1\%) & $0.1289\dagger$ (-2\%) & $0.1133\dagger$  (-2\%) & 0.1862    (+1\%)   & 0.2730$\dagger$ (+0\%) \\
    Human (Oracle) & 0.1343 & 0.1312$\dagger$ & 0.1154$\dagger$  & 0.1839 & $0.2725\dagger$ \\
    \bottomrule
    \end{tabular}
\vspace{-3mm}
\end{table*}

The comparison of our simulated user and the baselines on the document retrieval performance, before and after answering a clarifying question, is presented in Table \ref{tab:retrieval}.
The first row of the table shows the performance of the BM25 with only the initial query as input.
The following rows report the performance of BM25 with input composed of the initial query, clarifying question, and answers generated by each of the models.
Answers in the last row of the table are generated by humans and taken directly from ClariQ.

We observe that neither of the baseline models improves over the \textit{query-only} baseline, i.e., we can not reject the null hypothesis of equal performance for neither of the metrics.
This suggest that LSTM-seq2seq and Transformer-seq2seq do not yield useful and relevant answers to posed clarifying questions.
In most of the cases, the answers generated by the baselines even have a negative effect on the document retrieval performance, suggesting that the answers confuse the retrieval model.

On the other hand, we observe clear and significant performance increase of our simulated user over both the \textit{query-only} and seq2seq baselines.
The performance is significant by all of the metrics, except nDCG@1 and precision@1.
Similarly, oracle answers significantly outperform both baselines.
This confirms the finding previous research, suggesting the document retrieval performance increases with answers to clarifying questions as input \cite{aliannejadi2019asking}.

Interestingly, human-generated answers do not perform better than the answers generated by our model. 
This finding provides an answer to the research question RQ3, i.e, can the retrieval model benefit from the answers generated by \ourMethod.
The equal performance of the generated and human answers on this task is in line with previously described analyses on the \emph{usefulness} of the generated responses.
Together, the studies strongly support the possibility of substituting a real user with the user simulator for answering clarifying questions in conversational search.

\vspace{-1mm}
\subsection{Qualitative analysis}

\begin{table}[t]
    \centering
    \caption{Qualitative analysis of answers generated by user simulator \ourMethod.}
    \vspace{-4mm}
    \label{tbl:qual}
    \resizebox{\columnwidth}{!}{%
    \begin{tabular}{lll}
    \toprule
         Information need  & Clarifying question & Generated answer  \\
         \midrule
         \makecell[tl]{How do you treat severe\\ angular cheilitis}  & \makecell[tl]{are you looking for the \\ definition of angular cheilitis?} & \makecell[tl]{no i want to know\\ about treatment} \\
         \Xhline{0.2\arrayrulewidth}
         \makecell[tl]{What home remedies are \\there for angular cheilitis} & \makecell[tl]{are you looking for the \\definition of angular cheilitis?} & \makecell[tl]{no i am looking for\\ home remedies}\\
         \Xhline{0.2\arrayrulewidth}
         \makecell[tl]{Find a list of hobby stores\\ in Maryland} & \makecell[tl]{is this for a school\\ craft or play?} & \makecell[tl]{no i want to find a list\\ of hobby stores in maryland} \\
         \Xhline{0.2\arrayrulewidth}
         \makecell[tl]{What hobby stores \\carry trains} & \makecell[tl]{what location do you want\\ to find a hobby store in?} & \makecell[tl]{i want to know where i\\ can purchase trains} \\         
         \Xhline{0.2\arrayrulewidth}
         \makecell[tl]{What hobby stores \\carry trains} & \makecell[tl]{is this a hobby\\ for an adult?} & yes \\
         \bottomrule
         
    \end{tabular}
    }%
    \vspace{-5mm}
\end{table}

\partitle{Single-turn analysis.}
In this Section, we analyse several conversation samples of our user simulator with a hypothetical conversational search system.
Table \ref{tbl:qual} shows four interaction examples. 
The user simulator \ourMethod is initialised with the information need description text.
Given an initial query (omitted in the table for space), the conversational search system asks a clarifying question to elucidate \ourMethod's intent.
Then, \ourMethod generates the answer to the prompted question.
The information need and the questions for these examples are taken from ClariQ development set.
We note that most of the TREC-style datasets contain the information need (facet/topic) description alongside the initial query.
Thus, our simulated user can help in evaluation of conversational search systems on any of such datasets, as it only requires the information need description as initialisation. 
Then, the system we aim to evaluate can produce clarifying questions and receive answers from \ourMethod.

First two examples in the Table \ref{tbl:qual} initialise \ourMethod with different information needs. However, given the same initial query ``How to cure angular cheilitis'' and the same prompted clarifying question, \ourMethod answers differently, in line with the actual information need for each of the cases.
In the last three rows the table, we have different information needs for one broad topic of hobby stores.
Given the initial query ``I'm looking for information on hobby stores'', \ourMethod again answers questions in line with the underlying information need.
We notice that the text produced by our GPT-2-based user simulator is coherent and fluent, and, in given examples, indeed in line with the underlying information need.
Moreover, \ourMethod is not bound by answering the question in a ``yes'' or ``no'' fashion, but can rather produce various types of answers and even express its uncertainty (e.g., ``I don't know'').

\partitle{Multi-turn analysis.}
We perform initial case study on the multi-turn variant of \ourMethod.
While the initial analysis of multi-turn conversations suggests that \emph{usefulness} and \emph{naturalness} of single-turn interactions transfer into a multi-turn setting, additional evaluation is needed to strongly support that claim.
Thus, future work includes a pair-wise comparison of multi-turn conversations, inspired by ACUTE-Eval \cite{li2019acute}.

Moreover, we aim to observe user simulator behaviour in unexpected, edge case scenarios.
For example, initial analysis of the created multi-turn dataset showed that humans tend to repeat their previous answer when the clarifying question is off-topic or repeated. 
Similarly, our multi-turn \ourMethod has been observed to generate answers such as ``I already told you what I'm looking for'' when prompted with a repeated question.
However, such edge cases tend to confuse the multi-turn model, which leads to higher presence of hallucination than in the single-turn variation. 
This means that the user simulator drifts off the topic of the conversation and starts generating answers that are not not inline with the actual information need.
This effect is well-documented in recent literature on text generation \cite{dziri2021neural} and should be approached carefully.
Although edge cases are as well present in the acquired dataset, the GPT2-based model needs additional mechanisms in order to simulate the behaviour of users in such cases.
We leave deeper analysis of the topic for future research.

\vspace{-2mm}
\section{Conclusions}
In this paper, we have proposed a user simulator \ourMethod for alleviating evaluation of mixed-initiative conversational search systems.
More specifically, we demonstrated the feasibility of substituting expensive and time-consuming user studies with scalable and inexpensive user simulator.
Through a number of experiments, including automated metrics and crowdsourcing studies, we showed \ourMethod's capabilities in generating fluent and accurate answers to clarifying questions prompted by the search system.
In fact, a crowdsourcing study of answer \emph{usefulness} and \emph{naturalness} showed that answers generated by \ourMethod tied with human-generated answers in 51\% and 45\% of cases, respectively.
Moreover, we demonstrated the positive impact of generated answers on the performance of the retrieval model of the conversational search system, as the performance significantly increased when the answers to clarifying questions were taken into account.

Furthermore, we acquired an additional dataset for the training of the multi-turn model.
Specifically, we utilise crowdsourcing workers to gather multi-turn question-answer interaction about certain topics, where one worker takes the role of a search system and asks question, while the other worker responds to them.
Finally, we performed qualitative analysis of answers generated by \ourMethod.
We publicly release the code and the datasets for future research.

\partitle{Acknowledgements.}
This work was supported in part by
    the NWO Innovational Research Incentives Scheme Vidi (016.Vidi.189.039).

\bibliographystyle{ACM-Reference-Format}
\bibliography{mybib}



\end{document}